\documentclass{article}
\usepackage{ijcai21}

\usepackage{tikz}
\usepackage{amsmath}
\usepackage{amssymb}
\usepackage{subcaption}
\usepackage{booktabs}
\usepackage{tikzit}
\usepackage{multirow}

\tikzstyle{morphism}=[fill=white, draw=black, shape=rectangle, minimum width=1cm, minimum height=0.75cm]

\tikzstyle{functor}=[->, >=stealth]
\tikzstyle{dotted}=[-, dashed]
\tikzstyle{double arrow}=[<->, >=stealth]

\usepackage{times}
\usepackage{soul}
\usepackage{url}
\usepackage[hidelinks]{hyperref}
\usepackage[utf8]{inputenc}
\usepackage{graphicx}
\usepackage{amsmath}
\usepackage{amsthm}
\usepackage{booktabs}
\usepackage{algorithm}
\usepackage{algorithmic}
\title{Encoding Compositionality in Classical Planning Solutions}

\author{
Angeline Aguinaldo$^{1, 2}$
\and
William Regli$^1$\and
\affiliations
$^1$University of Maryland, College Park\\
$^2$Johns Hopkins University Applied Physics Laboratory\\
\emails
\{aaguinal, regli\}@umd.edu
}

\begin{document}

\maketitle

\begin{abstract}
Classical AI planners provide solutions to planning problems in the form of long and opaque text outputs. To aid in the understanding transferability of planning solutions, it is necessary to have a rich and comprehensible representation for both human and computers beyond the current line-by-line text notation. In particular, it is desirable to encode the trace of literals throughout the plan to capture the dependencies between actions selected. The approach of this paper is to view the actions as maps between literals and the selected plan as a composition of those maps. The mathematical theory, called category theory, provides the relevant structures for capturing maps, their compositions, and maps between compositions. We employ this theory to propose an algorithm agnostic, model-based representation for domains, problems, and plans expressed in the commonly used planning description language, PDDL. This category theoretic representation is accompanied by a graphical syntax in addition to a linear notation, similar to algebraic expressions, that can be used to infer literals used at every step of the plan. This provides the appropriate constructive abstraction and facilitates comprehension for human operators. In this paper, we demonstrate this on a plan within the Blocksworld domain.

\end{abstract}

\section{Introduction}
\label{sec:introduction}
Plans generated by automated planners for real-world problems are often long and opaque. They are multi-line text files, where every line describes the action and the associated parameters for a given step. This is simple and easy to parse, but does not encode other important information such as the trace of literals through the plan. We argue that a topological structure that encodes both the action sequence and the cause for that sequence, via the trace of literals, are necessary for determining whether the solution or a subsequence can be transferred, where transferring is seen as sequential composition of that with another plan in the same domain. A topological encoding alone, however, is not enough to aid human operators in identifying the skills and tasks worth transferring. The work of isolating the transferable tasks is offloaded to the human operator for now. This motivates the need for a representation that both encodes the provenance of literals with an emphasis on composition, and is comprehensible to human operators.



The widely-used convention for encoding classical artificial intelligence (AI) planning problems is the Problem Domain Definition Language (PDDL) \cite{Ghallab1998}. At the backbone of PDDL reasoning are logical propositions, or literals. These literals describe the preconditions and effects of an action. Provided a sequence of actions in a plan, it is likely that not all of an action's effects map to the next action's preconditions. This makes it difficult to understand which literals are informing each action because they may have been introduced many steps prior. When trying to understand whether the the plan or subsequence of it can be transferred, a user may inquire about the purpose each action has in achieving the goal in. To address this, this paper proposes using a structure, called string diagrams, that is defined in the branch of mathematics called category theory. Category theory is an algebraic system that is attentive to functions and how they compose. Validating compositionality between within and between plans is exactly the property that can be leveraged to evaluate the transferability of skills to other plans.

In this paper, we provide introductions to PDDL, category theory, and string diagrams and discuss related representations for PDDL domains and plans. We then present an example visualization for the Blocksworld domain file and then describe how we encoded PDDL domains, problems, and plans into the string diagram representation. Lastly, we discuss notable observations and how this representation can be extended to support explainable planning.

\section{Background}

\subsection{Planning Domain Definition Language (PDDL)}
\label{sec:pddl}
The prevalent classical planning language for describing domains and problems is the Problem Domain Definition Language (PDDL) \cite{Ghallab1998}. The schema adopted by PDDL at inception was based on the language used by the Stanford Research Institute Problem Solver (STRIPS) \cite{Fikes1971}. This requires a set of propositions, $F$, a set of operators with preconditions and effects, $O$, an initial state, $I$, and a goal state, $G$,  and operates under the \textit{closed world assumption}---all absent information is negative information. In other words, a STRIPS-based PDDL planning model can be defined as $P = <F, O, I, G>$. State models, $S(P)$, is a set of states, $s \in S(P)$, such that its elements are propositions. A transition function, $f: A \times S \rightarrow S$, maps between states according to the action applied. A corresponding cost is computed using a cost function, $\sigma: A \times S \rightarrow \mathbb{R}$. A plan, $\pi = <a_{i}, a_{i+1}, ... a_{n}>, a_{i} \in A$, is the sequence of actions that transition from the initial state, $s_0 = I$, to the goal state, $s_G = G$, according to the transition function, $f(a, s)$ and cost, $c(\pi) = \sum_{j=i}^{n} \sigma(a_j, s_j)$  \cite{Geffner2013}. There have been a number of solver heuristics and search algorithms developed that design cost functions according to soundness and optimality in order to identify plans efficiently \cite{Ghallab2004}.

\subsection{Model-based Representations for PDDL}
Explainable artificial intelligence (AI) planning (XAIP) is a subarea of research within the field of explainable AI (XAI) \cite{Gunning2017} whose goal is to relay to the user how and why a sequence of actions have been selected as a plan or policy \cite{Fox2017}. One approach to XAIP explanations is to use a representation that considers a plan in the context of the original domain model, i.e. \textit{model-based representation}. A model-based representation for AI plans is one that relies only on the solution, domain, and problem model provided by the user which means it is agnostic to the method used to produce the plan and is typically more relevant to the end user than algorithm-based or planning-based representations \cite{Chakraborti2020}. 

A common investigatory question that is asked of a model-based representation is \textit{"Why is this action in the plan?"}. A good explanation for this query is one that shows how the goal depends on the chosen actions of the plan \cite{Chakraborti2020}. There are few model-based representations that provide such explanations for PDDL plans. A prominent contender is Dovetail \cite{Magnaguagno2016}. Dovetail is a 2D graphical representation that uses jigsaw puzzle shapes an analogs for literal preconditions and effects. Every literal in the plan is assigned a vertical position, or row, in the visualization and every action in the plan is given a column. As a result, consecutive action pieces can fit together if their literals are compatible. This metaphor is intended to communicate the naturalness of the selected chain of actions. This is a creative view of the plan, however, it lacks the formal structure that would be used to provide information without visual inspection.

An example of mathematical model-based representations for PDDL are directed graphs. GIPO, Graphical Interface for Planning with Objects, \cite{Simpson2007} and VisPlan \cite{Glinsky2011} propose the use of directed graphs to model PDDL plans. Graph representations are useful for encoding relationships between actions according to their ingoing and outgoing literals, but their mathematical structures do not inherently capture the order in which actions are executed which is necessary for explaining causal dependency of actions.

\subsection{String Diagrams and Category Theory}
\label{sec:ct}
Eilenberg and MacLane \cite{Eilenberg1945} introduced the concepts of category theory in their study of algebraic topology as a way to transfer theorems between algebra and topology. In doing so, they provided a mathematical language that lifts many mathematical and non-mathematical concepts to this notion of maps between entities and compositions of those maps. This abstraction has found its usefulness in modeling natural language \cite{Coecke2010}, manufacturing processes \cite{Breiner2019}, database schema integration \cite{Shinavier2019}, biological protein structures \cite{Spivak2011}, and many other domains that require observing the interactions between entities, as opposed to the entities themselves. Likewise, these concepts provide a useful presentation of AI planning domains and plans because the resulting plans can be thought of as serial and parallel composition of maps between states.

To formally specify this representation, a mathematical structure must be defined. In category theory, the mathematical structure used is called a \textit{category}. To define a category, $\mathbb{C}$, it must have a set of \textit{objects} $\{A, B, C, \dots\}$ and a set of \textit{morphisms} $\{f, g, \dots\}$ that map between objects. The map for a given morphism can be written as $f: A \rightarrow B$. The source object is called the \textit{domain} and the target object is called the \textit{codomain} \cite{Spivak2014}. These objects and morphisms satisfy the following \cite{Lane1971}:


\begin{itemize}
\item For every object, there exists an \textit{identity morphism}.
\begin{equation}
    \forall  A \in \mathbb{C}, \; id_{A}: A \rightarrow A
\end{equation}
\item The \textit{composition} operation, $\circ$, acts on morphisms. Morphisms are composable when the codomain of a morphism exactly equals the domain of another morphism.
\begin{align*}
    f: A \rightarrow B, \;\;\; 
    g: B \rightarrow C 
\end{align*}
\vspace{-0.4cm}
\begin{equation}
    g \circ f: A \rightarrow C
\end{equation}
\item The composition of morphisms is \textit{associative}.
\begin{align*}
    f: A \rightarrow B, \;\;\;
    g: B \rightarrow C, \;\;\;
    h: C \rightarrow D
\end{align*}
\vspace{-0.4cm}
\begin{equation}
    (h \circ g) \circ f = h \circ (g \circ f)
\end{equation}
\item Identity morphisms act as a \textit{left} and \textit{right unitor of composition}. 
\begin{equation}
    id_{B} \circ f = f = f \circ id_{A}   
\end{equation}
\end{itemize}

These properties enforce consistent behavior for when more morphisms are composed together.

\begin{figure}[t!]
     \centering
     \begin{subfigure}[b]{0.2\textwidth}
         \centering
         \scalebox{0.7}{\begin{tikzpicture}
	\begin{pgfonlayer}{nodelayer}
		\node [style=none] (0) at (0, 3.25) {};
		\node [style=none] (1) at (0, 0.5) {};
		\node [style=none] (2) at (0, 3.75) {$A$};
		\node [style=none] (3) at (0, 0) {$B$};
		\node [style=none] (4) at (0.5, 1.75) {$f$};
		\node [style=none] (5) at (0, -0.5) {};
		\node [style=none] (6) at (0, -3.25) {};
		\node [style=none] (7) at (0, -3.75) {$C$};
		\node [style=none] (8) at (0.5, -2) {$g$};
	\end{pgfonlayer}
	\begin{pgfonlayer}{edgelayer}
		\draw [style=functor] (0.center) to (1.center);
		\draw [style=functor] (5.center) to (6.center);
	\end{pgfonlayer}
\end{tikzpicture}}
         \caption{Conventional syntax}
         \label{fig:composition}
     \end{subfigure}
     \centering
     \begin{subfigure}[b]{0.2\textwidth}
         \centering
         \scalebox{0.7}{\begin{tikzpicture}
	\begin{pgfonlayer}{nodelayer}
		\node [style=none] (0) at (0, 2.75) {};
		\node [style=none] (1) at (0, 0.75) {};
		\node [style=none] (2) at (0.5, 1.75) {$A$};
		\node [style=none] (3) at (0.5, -4.25) {$C$};
		\node [style=none] (6) at (0, -0.25) {};
		\node [style=none] (7) at (0, -2.25) {};
		\node [style=morphism] (9) at (0, 0.25) {$f$};
		\node [style=morphism] (10) at (0, -2.75) {$g$};
		\node [style=none] (11) at (0, -3.25) {};
		\node [style=none] (12) at (0, -5.25) {};
		\node [style=none] (13) at (0.5, -1.25) {$B$};
	\end{pgfonlayer}
	\begin{pgfonlayer}{edgelayer}
		\draw (0.center) to (1.center);
		\draw (6.center) to (7.center);
		\draw (11.center) to (12.center);
	\end{pgfonlayer}
\end{tikzpicture}}
         \caption{String diagram syntax}
         \label{fig:string_diagram}
     \end{subfigure}
     \caption{Two graphical representations for the composition of morphisms $f: A \rightarrow B$ and $g: B \rightarrow C$. In the case of (b), the lines can be called \textit{strings} and these strings symbolize identity morphisms, i.e. $id_A$. A common shorthand, however, is to simply label the string according to the object as shown in (b).}
     \label{fig:category}
 \end{figure}
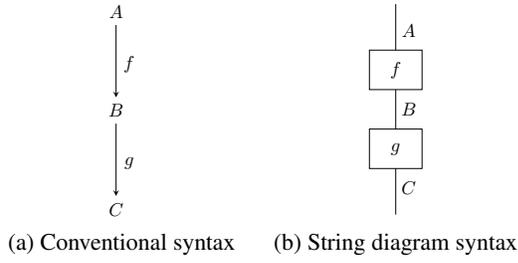

\begin{figure}[t!]
    \centering
     \begin{subfigure}[b]{0.2\textwidth}
         \centering
         \scalebox{0.6}{\begin{tikzpicture}
	\begin{pgfonlayer}{nodelayer}
		\node [style=none] (0) at (-2.75, 5) {};
		\node [style=none] (1) at (-0.25, 5) {};
		\node [style=none] (2) at (1.5, 5) {};
		\node [style=morphism] (3) at (-1.5, 3.25) {$B_{X, Y}$};
		\node [style=morphism] (4) at (0, -0.5) {$\alpha_{Y, X, Z}$};
		\node [style=none] (5) at (-0.75, 0) {};
		\node [style=none] (6) at (0, 0) {};
		\node [style=none] (7) at (2, 3.75) {};
		\node [style=none] (8) at (-2, 3.5) {};
		\node [style=none] (9) at (-1, 3.5) {};
		\node [style=none] (10) at (-2, 3) {};
		\node [style=none] (11) at (-1, 3) {};
		\node [style=none] (12) at (0, -1) {};
		\node [style=none] (13) at (-0.75, -1) {};
		\node [style=none] (14) at (3.5, 5) {};
		\node [style=none] (15) at (3, 3.75) {};
		\node [style=none] (16) at (0.75, -1) {};
		\node [style=none] (17) at (-2, -3.5) {};
		\node [style=none] (18) at (0, -3.5) {};
		\node [style=none] (19) at (2, -3.5) {};
		\node [style=morphism] (20) at (2.5, 3.25) {$\rho$};
		\node [style=none] (21) at (2.5, 2.75) {};
		\node [style=none] (22) at (0.75, 0) {};
		\node [style=none] (23) at (-2.75, 5.5) {$X$};
		\node [style=none] (24) at (-0.25, 5.5) {$Y$};
		\node [style=none] (25) at (1.5, 5.5) {$Z$};
		\node [style=none] (26) at (3.5, 5.5) {$I$};
		\node [style=none] (27) at (2, 1.25) {$Z$};
		\node [style=none] (28) at (0.25, 1.25) {$X$};
		\node [style=none] (29) at (-1.75, 1.25) {$Y$};
		\node [style=none] (30) at (-2, -4) {$Y$};
		\node [style=none] (31) at (0, -4) {$X$};
		\node [style=none] (32) at (2, -4) {$Z$};
	\end{pgfonlayer}
	\begin{pgfonlayer}{edgelayer}
		\draw [in=90, out=-90] (0.center) to (8.center);
		\draw [in=90, out=-90] (1.center) to (9.center);
		\draw [in=90, out=-90] (10.center) to (5.center);
		\draw [in=90, out=-90, looseness=1.25] (11.center) to (6.center);
		\draw [in=90, out=-90] (2.center) to (7.center);
		\draw [in=90, out=-90] (13.center) to (17.center);
		\draw [in=90, out=-90] (12.center) to (18.center);
		\draw [in=90, out=-90, looseness=0.75] (16.center) to (19.center);
		\draw [in=90, out=-75] (14.center) to (15.center);
		\draw [in=90, out=-90, looseness=0.75] (21.center) to (22.center);
	\end{pgfonlayer}
\end{tikzpicture}}
         \caption{Explicit}
         \label{fig:smc_explicit}
     \end{subfigure}
     \centering
     \begin{subfigure}[b]{0.2\textwidth}
         \centering
         \scalebox{0.7}{\begin{tikzpicture}
	\begin{pgfonlayer}{nodelayer}
		\node [style=morphism] (4) at (0, 0) {$\alpha_{Y, X, Z}$};
		\node [style=none] (5) at (-0.75, 0.5) {};
		\node [style=none] (6) at (0, 0.5) {};
		\node [style=none] (10) at (0, 3.25) {};
		\node [style=none] (11) at (-2, 3.25) {};
		\node [style=none] (12) at (0, -0.5) {};
		\node [style=none] (13) at (-0.75, -0.5) {};
		\node [style=none] (16) at (0.75, -0.5) {};
		\node [style=none] (17) at (-2, -2.75) {};
		\node [style=none] (18) at (0, -2.75) {};
		\node [style=none] (19) at (2, -2.75) {};
		\node [style=none] (21) at (2, 3.25) {};
		\node [style=none] (22) at (0.75, 0.5) {};
		\node [style=none] (23) at (-2, 3.75) {$X$};
		\node [style=none] (24) at (0, 3.75) {$Y$};
		\node [style=none] (25) at (2, 3.75) {$Z$};
		\node [style=none] (30) at (-2, -3.25) {$Y$};
		\node [style=none] (31) at (0, -3.25) {$X$};
		\node [style=none] (32) at (2, -3.25) {$Z$};
	\end{pgfonlayer}
	\begin{pgfonlayer}{edgelayer}
		\draw [in=90, out=-90] (10.center) to (5.center);
		\draw [in=90, out=-90, looseness=1.25] (11.center) to (6.center);
		\draw [in=90, out=-90] (13.center) to (17.center);
		\draw [in=90, out=-90] (12.center) to (18.center);
		\draw [in=90, out=-90, looseness=0.75] (16.center) to (19.center);
		\draw [in=90, out=-90, looseness=0.75] (21.center) to (22.center);
	\end{pgfonlayer}
\end{tikzpicture}}
         \caption{Shorthand}
         \label{fig:smc_shorthand}
     \end{subfigure}
     \caption{A string diagram representation of the symmetric monoidal category properties shown in Equations 5-7. It is typical convention in string diagrams to replace rectangles symbolizing braids, such as $B_{X, Y}$ in (a), with crossing strings as seen in (b); and a tensor product with identity objects as just the non-identity objects.}
     \label{fig:smc}
 \end{figure}
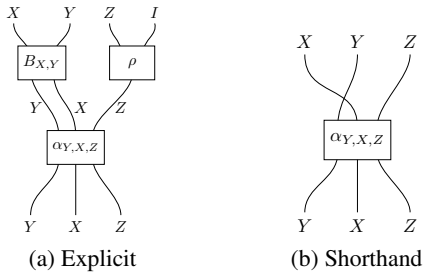

To model more complex maps, it is necessary that the structure support multiple objects in the domain and codomain. Additional mathematical structure (tensor product $\otimes$) can be added to the definition of a category to support this. This enhanced category definition is known as a \textit{symmetric monoidal category} \cite{Joyal1991}, $\mathbb{M}$, and has the following additional properties:

\begin{itemize}
\item A \textit{unit object}, $I \in \mathbb{M}$
\item A map, called the \textit{tensor product} $\otimes$, which is the product of $\mathbb{M}$ with itself,  $\otimes: \mathbb{M} \times \mathbb{M} \rightarrow \mathbb{M}$.
\item This tensor product is \textit{associative},
    \begin{equation}
        a_{X, Y, Z} : (X \otimes Y) \otimes Z \rightarrow X \otimes (Y \otimes Z) 
    \end{equation}
\item has \textit{left} and \textit{right unitor isomorphisms},
    \begin{equation}
        \rho_l: I \otimes X \rightarrow X \;\;\;\;\;\;
        \rho_r: X \otimes I \rightarrow X
    \end{equation}
\item and has a \textit{braiding isomorphism} that is \textit{symmetric} \cite{Joyal1991}
    \begin{equation}
        B_{X,Y}: X \otimes Y \rightarrow Y \otimes X
    \end{equation}
\end{itemize}

These properties allow for both parallel and serial composition of processes. This also enables partial composition, where only a subset of domain strings of a morphism match the codomain strings of another morphism.  

In addition to this structure, category theory provides a graphical syntax for illustrating maps and their compositions. Figure \ref{fig:category} shows the composition of $g \circ f: A \rightarrow C$ from Equation 2 in conventional syntax and a Poincare dual syntax. The Poincare dual syntax shows the morphisms $f$ and $g$ as rectangles and $A, B, C$ objects as lines. Figure \ref{fig:string_diagram} can be extended to support the symmetric monoidal category structure as seen in Figure \ref{fig:smc}, where multiple lines can pass through each rectangle. These diagrams are known as \textit{string diagrams}. 

\begin{figure}[t]
    \centering
    \includegraphics[width=0.27\textwidth]{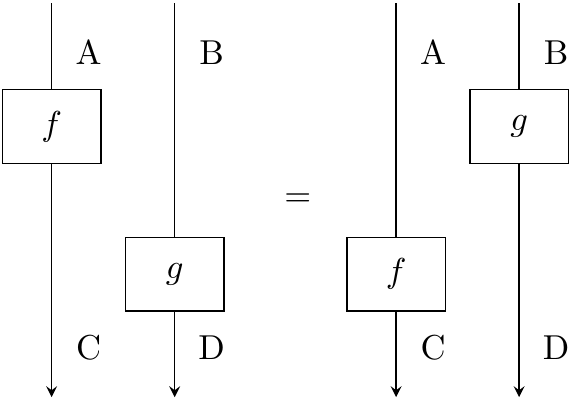}
    \caption{An example of deformation invariance in string diagrams. This feature allows the sliding of rectangles along strings in a planar fashion in accordance with the composition operator.}
    \label{fig:deformation}
\end{figure}

In this syntax, horizontally adjacent strings represent the tensor product of those objects, i.e. $X \otimes Y$, which is another object in $\mathbb{M}$. The identity morphism of that object, i.e. $id_{X \otimes Y}$, can be used when composing, $\circ$, with other morphisms. An expression that describes the composition, $\circ$, and tensor product, $\otimes$, of morphisms is the linear notation that encodes its graphical representation. For example, the linear notation for Figure \ref{fig:smc_explicit} can be seen in Equation \ref{eq:linear}.

\begin{equation}
\begin{aligned}
    (id_{X} \otimes id_{Y} \otimes id_{Z} \otimes id_{I})
    \circ \\ (B_{X,Y} \otimes \rho)
    \circ \\ (id_{Y} \otimes id_{X} \otimes id_{Z})
    \circ \\ (\alpha_{Y, X, Z})
    \circ \\ (id_{Y} \otimes id_{X} \otimes id_{Z})
\end{aligned}
\label{eq:linear}
\end{equation}
The morphisms and the order in which they are composed can be seen clearly from this notation.

A noteworthy consequence of the tensor product and its properties is the equivalence between strings diagrams under planar deformation \cite{Selinger2010}. For example, we can consider the equivalence found in Equation 9 and its corresponding graphical syntax shown in Figure \ref{fig:deformation}:
\begin{equation}
    (id_C \otimes g) \circ (f \otimes id_B) = (f \otimes id_D) \circ (id_A \otimes g)
\end{equation}
This gives the appearance of sliding rectangles past each other along strings which may be useful when seeking different sequences of compositions.

\section{String Diagram Notation for PDDL}
\label{sec:string_diagram_notation}
In this section, we describe how we have chosen to translate key elements of PDDL domain, problem, and plan descriptions to mathematical structures from category theory. We construct two categories, one for the domain file, $\mathbb{D}$, and another of the problem file, $\mathbb{P}$. Table \ref{tab:correspondence} summarizes this correspondence. Currently, this notation only supports the STRIPS requirement of PDDL 1.2 \cite{Ghallab1998}. Terms provided by other extensions such as typing, equality, condition-effects are not handled.

\subsection{Domain File}
The domain file provides the domain and codomain signatures for morphisms in our category. STRIPS-based domain files consists of the domain name, predicates, and actions. 

\paragraph{Actions}
Actions are operator data models that describe the parameters, preconditions, and effects of the given action, denoted by the \texttt{:action} token. Preconditions and effects values are conjunctions ($\wedge$) of logical predicates. A conjunction is distinguished by a pair of parentheses and the term \texttt{and}. The term \texttt{not} can prefix a predicate. When storing this negation, the parser prefixes the predicate name with the negation symbol, $\neg$. In the string diagram representation, actions are identified as morphisms, where each predicate in the precondition is a domain string and each predicate in the effect is a codomain string.



\paragraph{Predicates}
Predicates are terms that serve as placeholders for data with logical states. Predicates are referenced in an action's preconditions and effects. They typically refer to aspects of the domain whose status will influence planning decisions. Some example predicates from the Blocksworld domain file include \textit{(on ?x ?y)}, \textit{(holding ?x)}, \textit{(handempty)}. Predicates can be identified as strings when representing actions in the domain file as string diagrams. Provided a problem file and plan, parameters in the predicate signatures can be replaced with PDDL objects. 

\paragraph{Parameters}
Parameters for both the predicates and actions are prefixed with \texttt{?}. When parameters are used in predicate definitions, the characters following each \texttt{?} is used to distinguish that parameter from other parameters associated with that predicate. For example, in the predicate \textit{(on ?x ?y)}, the characters \textit{x} and \textit{y} are not required symbols when using the predicate in the action conditions. In effect, these parameters are used solely to specify the arity of the predicate. These parameters will be substituted with objects specified by a PDDL plan. Parameters do not directly map to a structure in the string diagram representation.

\subsection{Problem File}
The problem file informs the initial and goal strings in the string diagram. STRIPS-based problem files consist of the problem name, domain name, objects, an initial state, and a goal state. 

\paragraph{Objects}
Objects are symbols that are used to populate parameters, denoted by the \texttt{:objects} token. Objects do not directly map to a structure in the string diagram representation.

\paragraph{Literals}
When predicate parameters are populated, according to a PDDL solver, they become literals, and are used as data for preconditions and effects in actions. In a string diagram representation, every uniquely parameterized predicate is treated as a string. For example, if two predicates \textit{(ontable A)} and \textit{(ontable B)} are constructed according to a PDDL plan, they are unique strings. Notice that this also implies that negated versions of literals are considered distinct from their positive selves. For example, \textit{(ontable A)} and \textit{(not (ontable A))} are distinct strings and their relationship by logical negation is not encoded.


\paragraph{Initial State}
The initial state, denoted by \texttt{:init} token in the problem file, is a conjunction of literal assumptions that are true at the beginning of the planning problem. All literals in the initial state are represented by their own strings. These strings are tensored, $\otimes$, together and serve as the first morphism in the chain of compositions. In a later step, additional literals may be tensored with the initial state. This happens when an action in the plan makes an assumption about the initial state that is not explicitly stated in the problem definition.

\paragraph{Goal State}
The goal state, denoted by \texttt{:goal} token, is a list of literals that \textit{must} be true at the end of the planning problem. All literals in the goal state are represented by their own strings. These strings are $\otimes$ together and serve as the last morphism in the chain of compositions. In a later step, additional literals may be tensored with the goal state if an action in the plan has an effect that is not explicitly required by the goal definition.

\begin{table}[]
\centering
\begin{tabular}{@{}llll@{}}
\toprule
PDDL File & Description & Category Theory & String Diagram\\ \midrule
\multirow{3}{*}{Domain, $\mathbb{D}$} & Actions & Morphisms & Rectangles \\ \cmidrule(l){2-4} 
 & Predicates & Objects & Strings\\ \cmidrule(l){2-4} 
 & Parameters & -- \\ \midrule
\multirow{4}{*}{Problem, $\mathbb{P}$} & Objects & -- & -- \\ \cmidrule(l){2-4} 
 & Literals & Objects & Strings\\ \cmidrule(l){2-4} 
 & Initial State & Morphism & Rectangles\\ \cmidrule(l){2-4}
 & Goal State & Morphism & Rectangles\\ \bottomrule
\end{tabular}
\caption{Correspondence between category theory structures and description components found in PDDL domain and problem files}
\label{tab:correspondence}
\end{table}

\subsection{Plan}
After the sequence of parameterized actions have been converted into morphisms, we chain, or compose, these morphisms to generate a fully connected diagram. A naive chaining algorithm was implemented to infer valid tensor products and compositions in scenarios where morphisms can only partially compose. The primary goal of this algorithm is to enforce the compatibility of domains and codomains of preceding and subsequent morphisms during composition.

In this algorithm, there exists a notion of horizontal \textit{slices} where each slice is a list of morphisms, including identities and braids, that will be tensored together from left to right. It is essential that the domain and codomain of the slices match in order to permit a valid composition. 

There are four main steps to the algorithm:

\begin{enumerate}
    \item \textit{Backward pass} to weave input strings of each morphism, from goal state to initial state (bottom-up).
    \item \textit{Forward pass} to weave output strings of each morphism, from initial state to goal state (top-down).
    \item \textit{Add braids} to add braids, or string swaps, in case the order of the tensor product is not compatible for composition.
    \item \textit{Compose} to chain the blocks from top-down. This constructs the string diagram.
\end{enumerate}

A byproduct of this chaining algorithm is the propagation of literals upstream and downstream which exposes those literals that are implicitly instantiated according to the plan.

\section{Example}
\label{sec:example}

A program written in the Julia programming language\footnote{https://julialang.org/} was developed in order to automatically generate string diagrams. This program parses the PDDL domain, problem, and solution files and encodes its elements into the symmetric monoidal category constructs, which are JSON-serializable.  Catlab\footnote{\url{https://github.com/epatters/Catlab.jl}}, a Julia-based category theory library, was leveraged for its constructors. These diagrams represent a subset of objects and morphisms in $\mathbb{P}$, also called a subcategory of $\mathbb{P}$. Currently, this program only supports PDDL files formatted using the STRIPS requirements. 



\begin{figure*}
     \centering
     \begin{subfigure}[b]{0.24\textwidth}
         \centering
         \includegraphics[width=\textwidth]{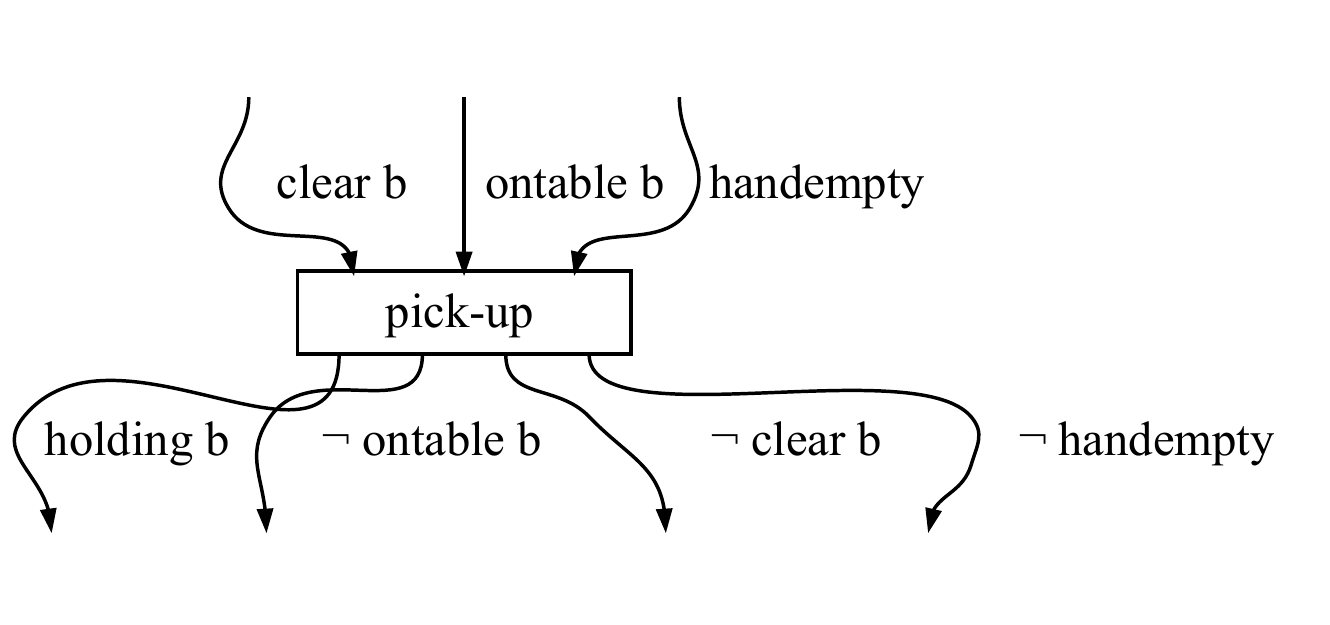}
         \caption{First step}
         \label{fig:step_0}
     \end{subfigure}
     \hfill
     \begin{subfigure}[b]{0.24\textwidth}
         \centering
         \includegraphics[width=\textwidth]{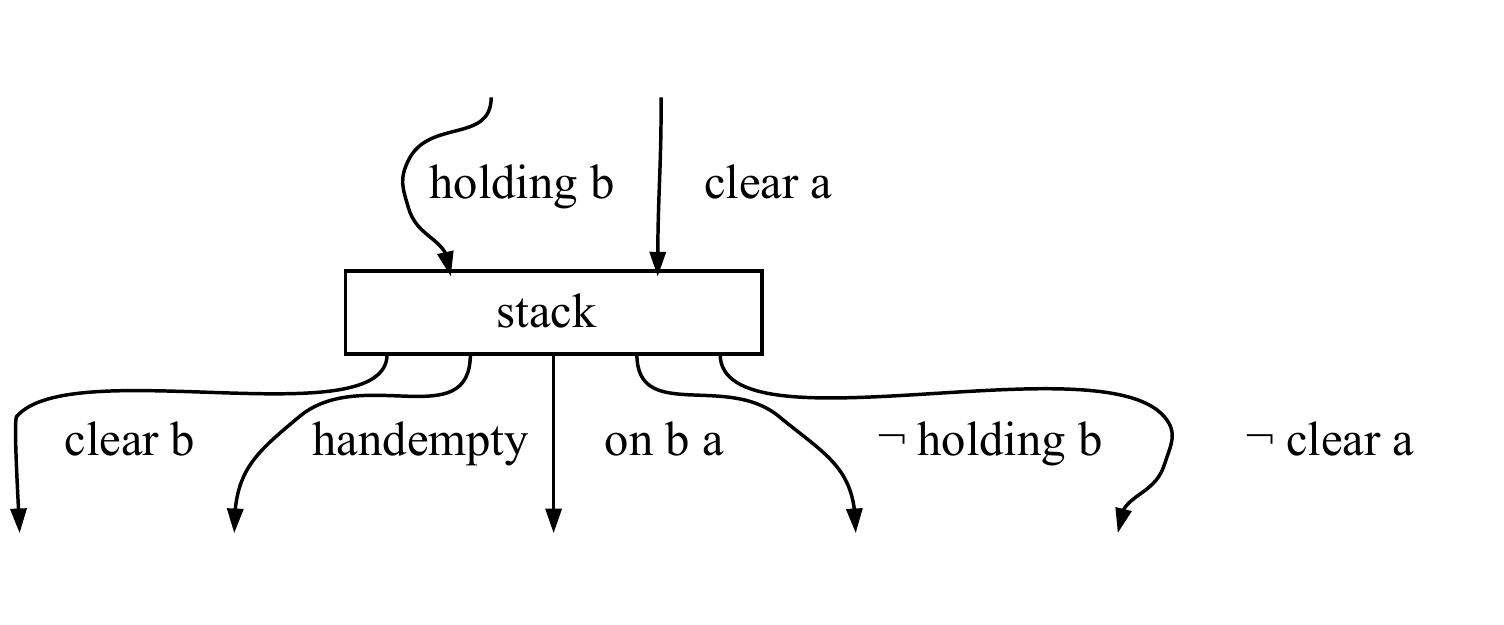}
         \caption{Second step}
         \label{fig:step_1}
     \end{subfigure}
     \hfill
     \begin{subfigure}[b]{0.24\textwidth}
         \centering
         \includegraphics[width=\textwidth]{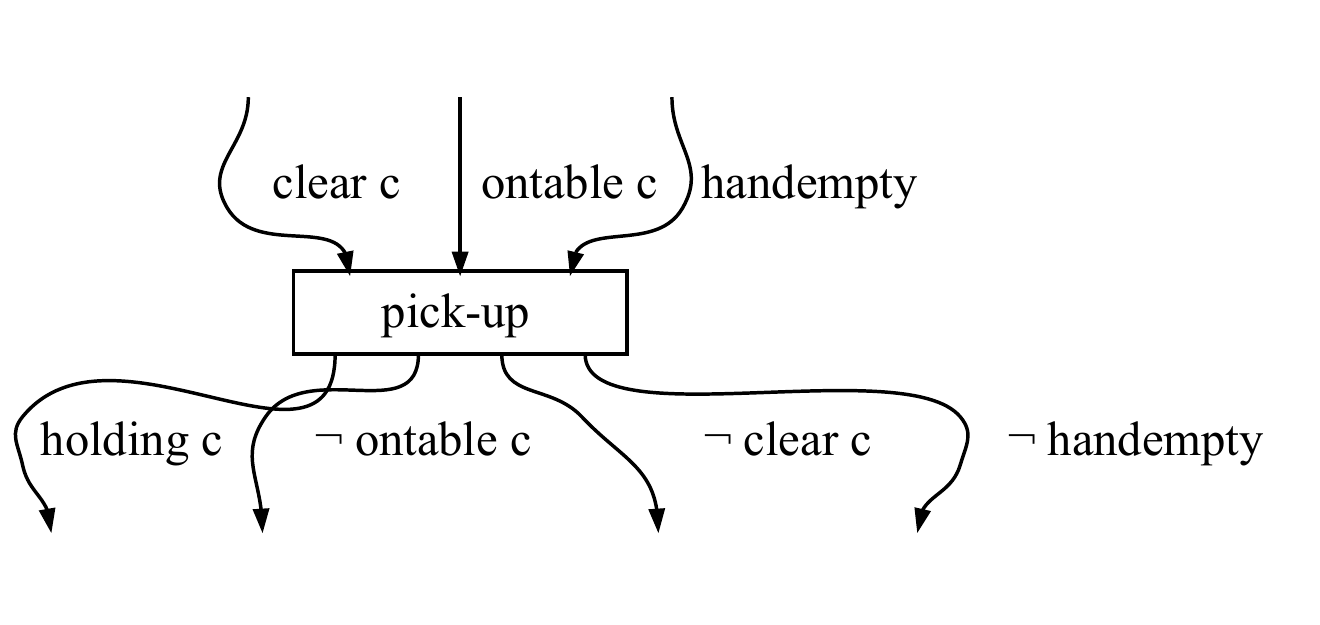}
         \caption{Third step}
         \label{fig:step_2}
     \end{subfigure}
     \hfill
     \begin{subfigure}[b]{0.24\textwidth}
         \centering
         \includegraphics[width=\textwidth]{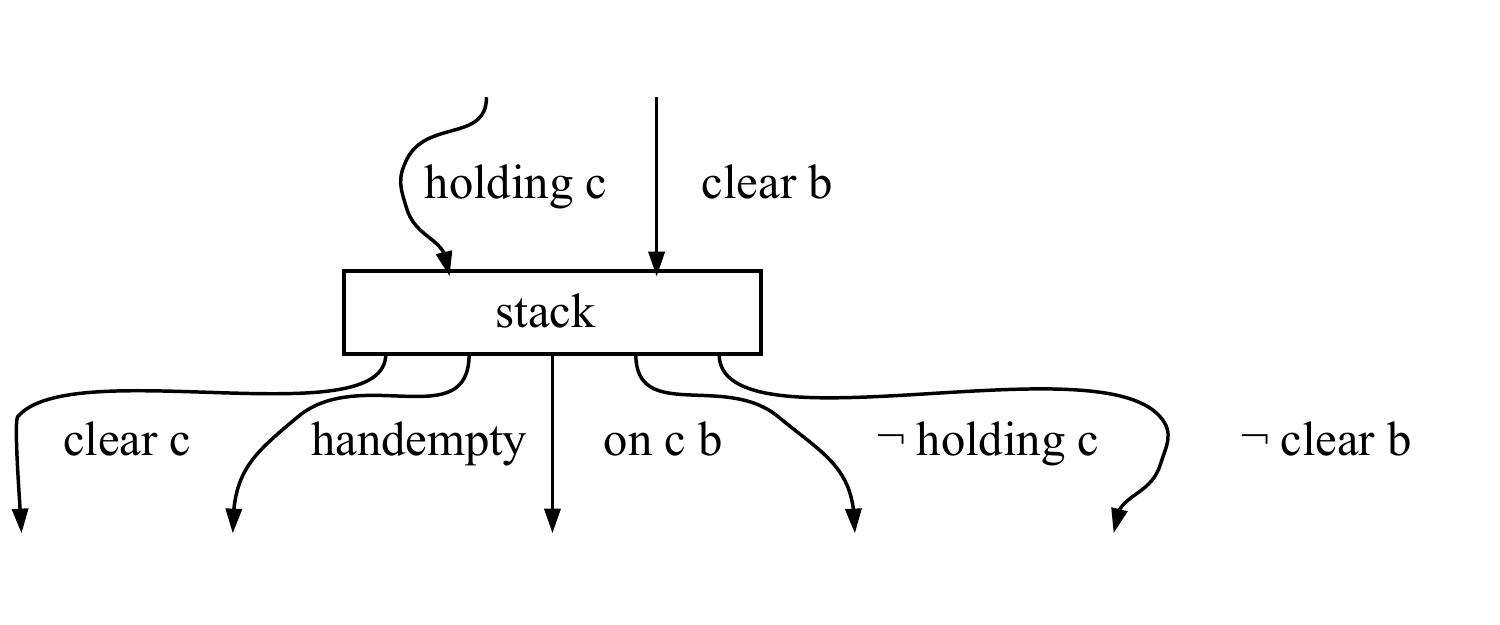}
         \caption{Fourth step}
         \label{fig:step_3}
     \end{subfigure}
        \caption{These are the morphisms corresponding to steps proposed by a PDDL solver for the Blocksworld domain and problem described in \textit{Examples} section. The diagrammatic description of the steps clearly depicts how the parameters populate the pre- and post-conditions to form literals; therefore, intuitively conveying the requirements of each action in the context of the problem.}
        \label{fig:plan_morphisms}
\end{figure*}

\subsection{\texttt{Blocksworld} domain}
The Blocksworld\footnote{https://github.com/pellierd/pddl4j/tree/master/pddl/blocksworld} domain describes a scenario where there are cube-shaped objects, called \textit{blocks}, on a table, denoted by the predicate (\textit{ontable ?x}). The objective is to stack the blocks according to the stacking configuration described by the goal. Only one block can fit on top of another block, (\textit{on ?x ?y}), which implies a block cannot be stacked on more than one block simulataneously. If a block is not underneath another block, it is considered clear, (\textit{clear ?x}). A hand is used to manipulate the configuration of the blocks, and can be empty (\textit{handempty}) or holding a block (\textit{holding ?x}).

The domain consists of four actions: pick up, put down, stack, unstack. 
\begin{itemize}
    \item \texttt{pick-up}: This action expects that a given block is clear, on the table, and the hand is empty. This action maps the state of the world such that the block is not on the table, the block is not clear, the hand is not empty, and the hand is holding the block. 
    \item \texttt{put-down}: This action expects that the hand is holding a block. This action maps the state of the world such that the hand is not holding the block, the hand is empty, the block is clear, and the block is on the table.
    
    \item \texttt{stack}: This action expects that the hand is holding a block (\texttt{?x}) and that the other block (\texttt{?y}) is clear. This action maps the state of the world such that the hand is not holding the block (\texttt{?x}), the other block (\texttt{?y}) is not clear, the hand is empty, and the block (\texttt{?x}) is on top of the other block (\texttt{?y}).
    
    \item \texttt{unstack}: This action expects that the block (\texttt{?x}) is on top of another block (\texttt{?y}), that the block (\texttt{?x}) is clear, and the hand is empty. This action maps the state of the world such that the hand is holding the block (\texttt{?x}), the other block (\texttt{?y}) is clear, the block (\texttt{?x}) is not clear, the hand is not empty, and the block (\texttt{?x}) is not on top of the other block (\texttt{?y}).
\end{itemize}
The problem in this example initialized three blocks as objects, (\textit{a, b, c}), and stated that all the blocks are clear and on the table, and that the hand is empty. The goal was to have block \textit{c} on top of \textit{b}, and block \textit{b} on top of \textit{a}.

\begin{verbatim}
(define (problem BLOCKS-3-0)
    (:domain BLOCKS)
    (:objects a b c)
    (:init (clear c) (clear a) 
            (clear b) (ontable c) 
            (ontable a) (ontable b) 
            (handempty))
    (:goal (AND (on c b) (on b a))))
\end{verbatim}

The PDDL4J toolkit\footnote{The default configurations of PDDL4J toolkit uses the heuristic search planner (HSP) \cite{Bonet2001} and the FF heuristic \cite{Hoffmann2001}.} \cite{Pellier2018} was run to determine the actions and parameters needed to transition the world from the initial state to the goal state. The planner output was parsed to extract the plan and ignore other outputs, such as cost and runtime. The parsed plan for this example can be seen below. 

\begin{verbatim}
pick-up b
stack b a
pick-up c
stack c b
\end{verbatim}

\begin{figure*}[h!]
    \centering
    \includegraphics[width=\textwidth]{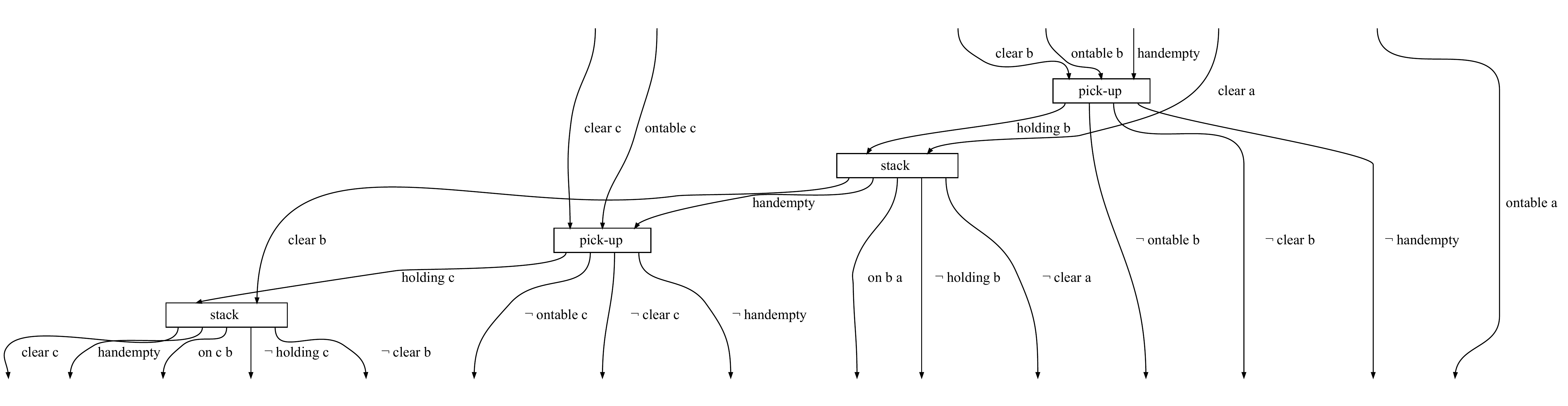}
    \caption{String diagram for the Blocksworld domain and problem. The diagram is read top-down. The curved lines in the illustration, i.e. strings, represent literals that are initialized or constructed according to the PDDL plan. The rectangles represent the actions from the domain that are referenced in the plan. The order in which the actions appear in the plan can be seen by following the strings from the top to the bottom and noting the order of the rectangles you encounter along the way. The entire diagram represents a map from initial state to goal state.}
    \label{fig:block_string_diagram}
\end{figure*}

The individual steps in the plan can be understood as morphisms, and likewise can be represented as string diagrams. An example is shown in Figure \ref{fig:plan_morphisms}. The diagrammatic description clearly depicts how the parameters populate the preconditions and effects to form literals; therefore, intuitively conveying the requirements of each action in the context of the problem. The fully-composed string diagram containing all the steps of the Blocksworld plan can be seen in Figure \ref{fig:block_string_diagram}. The top of the diagram shows strings representing the literals of the initial state and assumptions about the world that were introduced by the plan. The bottom of the diagram shows strings representing the literals of the goal state as well as the literals produced from the plan. The diagram is read from top to bottom and the actions of the plan appear as labeled rectangles. Table \ref{table:resources} shows a summary of all the satisfied literals at every step in the plan. This information can be derived from the linear notation that encodes the graphical representation.

\begin{figure}[t]
    \centering
    \includegraphics[width=0.35\textwidth]{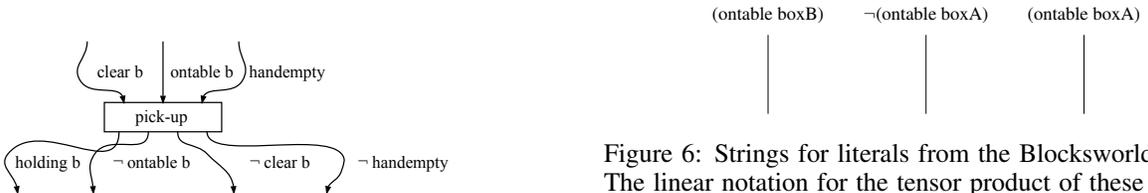}
    \caption{This an example morphism corresponding to the first action chosen by a PDDL solver for the Blocksworld domain. The linear notation for this action is $pick\_up: (id_{\text{(clear b)}} \otimes id_{\text{(ontable b)}} \otimes id_{\text{(handempty)}}) \rightarrow (id_{\text{(holding b)}} \otimes id_{\neg\text{(ontable b)}} \otimes id_{\neg\text{(clear b)}} \otimes id_{\neg\text{(handempty)}})$.}
    \label{fig:plan_morphisms}
\end{figure}

\begin{figure}[t]
    \centering
    \scalebox{0.7}{
     \begin{tikzpicture}
	\begin{pgfonlayer}{nodelayer}
		\node [style=none] (0) at (0, 2) {};
		\node [style=none] (1) at (0, -1) {};
		\node [style=none] (2) at (-6, 2) {};
		\node [style=none] (3) at (-6, -1) {};
		\node [style=none] (4) at (6, 2) {};
		\node [style=none] (5) at (6, -1) {};
		\node [style=none] (6) at (-6, 2.75) {(ontable boxB)};
		\node [style=none] (7) at (0, 2.75) {$\neg$(ontable boxA)};
		\node [style=none] (8) at (6, 2.75) {(ontable boxA)};
	\end{pgfonlayer}
	\begin{pgfonlayer}{edgelayer}
		\draw (2.center) to (3.center);
		\draw (0.center) to (1.center);
		\draw (4.center) to (5.center);
	\end{pgfonlayer}
\end{tikzpicture}
     }
     \caption{Strings for literals from the Blocksworld domain. The linear notation for the tensor product of these strings is $id_{\text{(ontable boxB)}} \otimes id_{\neg\text{(ontable boxA)}} \otimes id_{\text{(ontable boxA)}}$.}
     \label{fig:strings}
\end{figure}

\begin{table*}[h!]
\centering
\begin{tabular}{@{}cll@{}}
\toprule
\textit{\textbf{State}} & \multicolumn{1}{c}{\textit{\textbf{Satisfied Literals}}} & \textit{\textbf{Action}} \\ \midrule

\textbf{Initial} & \textbf{clear c}, \textbf{ontable c}, \textbf{clear b}, \textbf{ontable b}, \textbf{handempty}, \textbf{clear a}, \textbf{ontable a} &  \\ \midrule

\textbf{Step 1} & clear c, ontable c, \textbf{clear b}, \textbf{ontable b}, \textbf{handempty}, clear a, ontable a & pick-up \\ \midrule

\textbf{Step 2} & clear c, ontable c, \textbf{holding b}, \textbf{clear a},  $\neg$ontable b,$\neg$clear b, $\neg$handempty, ontable a & stack \\ \midrule

\textbf{Step 3} & \begin{tabular}[c]{@{}l@{}} clear b, \textbf{ clear c}, \textbf{ontable c}, \textbf{handempty}, on b a, $\neg$holding b, $\neg$clear a, $\neg$ontable b, $\neg$clear b, $\neg$handempty, \\ ontable a \end{tabular}& pick-up \\ \midrule

\textbf{Step 4} & \begin{tabular}[c]{@{}l@{}}\textbf{holding c}, \textbf{clear b}, $\neg$ontable c, $\neg$clear c, $\neg$handempty, on b a, $\neg$holding b, $\neg$clear a, $\neg$ontable b, \\ $\neg$clear b, $\neg$handempty, ontable a\end{tabular} & stack \\ \midrule

\textbf{Goal} & \begin{tabular}[c]{@{}l@{}}clear c, handempty, \textbf{on c b}, $\neg$holding c, $\neg$clear b, $\neg$ontable c$\neg$clear c, $\neg$handempty, \textbf{on b a}, $\neg$holding b, \\ $\neg$clear a, $\neg$ontable b, $\neg$clear b, $\neg$handempty, ontable a\end{tabular} &  \\ \bottomrule
\end{tabular}
\caption{The satisfied literals at every step of the plan can be concluded from the linear notation that encodes the string diagram representation shown in Figure \ref{fig:block_string_diagram}. You can check this by reading the diagram from left to right at varying horizontal positions and noting the strings you encounter. The explicitly required literals are highlighted in \textbf{bold text}. The effects for each action are included as \textit{Satisfied Literals} in the next row.}
\label{table:resources}
\end{table*}

\section{Discussion}

In this paper, we presented a model-based representation that describes a PDDL plan in the context of the domain description. From the string diagram depiction of the plan, we can clearly see what predicates were initialized, how they were used, and what predicates are present at any given point in the plan. Some notable observations that explicitly answer the question of \textit{"Why is this action in the plan?"} are listed here:
\begin{itemize}
    \item Following Figure \ref{fig:block_string_diagram} from top to bottom, we can observe that the first \textit{stack} relies on \textit{(holding b)} which is an effect from \textit{pick-up}. This implies that \textit{stack} depends on \textit{pick-up} in order to execute.
    \item Additionally, we can observe when each literal in the initial state is used to inform the action. For example, the \textit{(clear c)} and \textit{(ontable c)} literals do not get referenced until the second \textit{pick-up} action is called.
    \item From the \textbf{Start} row of Table \ref{table:resources}, we can see that no new literals were introduced as implicit assumptions according to the given plan. This is evidenced by the lack of unbolded literals. 
\end{itemize}

This representation shows the implicit changes occurring in the world that are not clearly evidenced by the simple line-by-line description of the plan.  This also exposes the unanticipated literals that could result in errors when operating under the closed world assumption. 

To consider whether we can compose this solution with another in the same domain, we simply need to evaluate the equality of domain of the current solution with the target codomain of the target solution or vice versa. If a desired subsequence of actions forms a skill, we simply need to select the morphisms we would like to compose and compare the domain or codomains of the specific composition with the target solution's domain and codomain. 

\subsection{Limitations}
The current encoding scheme presents some limitations. As aforementioned, the PDDL extensions supported are restricted. We currently do not have a way to visualize notation such as quantifiers, equalities, and other extensions. We are also unable to encode relationships between positive and negated version of the same literal, which are currently treated as independent information under the closed world assumption. Lastly, while the structure can still be used for encoding long plans with many actions, the visualization does not scale effectively. To handle this, it may be possible to design a heuristic to detect repeated patterns in actions so that the plan can be grouped into subtasks. 

\subsection{Benefits of category theory structure}

Recall that this representation hinges on structures defined in category theory, such as morphisms and objects. Morphisms are made by the actions described in PDDL domain files. These morphisms can be seen as unit string diagrams that can be composed, as shown in Figures \ref{fig:plan_morphisms}. This representation also provides the user with additional context for the actions, such as how the parameters populate the preconditions and effects. For example, in Figure \ref{fig:plan_morphisms}, the step \textit{pick-up b} is clearly depicted as having \textit{(clear b)}, \textit{(ontable b)}, \textit{(handempty)} as pre-conditions and \textit{(holding b)}, \textit{(not (ontable b))}, \textit{(not (clear b))}, \textit{(not (handempty))} as effects, which is not easily inferred by the text description of the plan.


We also introduced that fact that these morphisms can be vertically, $\circ$, and horizontally, $\otimes$, composed together. The linear expression translates to the layout of the rectangles and strings in the string diagrams. Evidently, the layout is particularly useful for conveying a sense of order and time in the plan. Another key aspect of this is that the string diagram representation is deformation invariant \cite{Selinger2010}, which means that sliding rectangles along strings is similar to rearranging terms of an algebraic equation. This implies that alternate but valid plans can be observed by reordering actions, i.e. rectangles. Notably, this structure is not limited in its application to PDDL, but can be applied to any declarative planning language. 

\subsection{Future Work}
Category theory, though not extensively described, permits defining maps between categories, whose construction is called a \textit{functor}. One way this can be used is to provide correspondence between $\mathbb{P}$ to a semantic category, such as one describing human conceptual models of planning tasks. Mapping multiple solutions, regardless of domain, to these conceptual models would provide a framework for documenting abstract concepts and identifying equivalent plans up to isomorphism. This exposition is left as future work. 
In addition, the graphical representation and formal linear notation provide opportunities for explanatory insights and intuitive visualizations of PDDL plans. For visualization, it is easy to imagine interactions such as highlighting the strings of a particular literal in order to witness its path through the plan, or sliding rectangles along strings to view alternative plans. 


\section{Conclusion}

This representation has the primary benefit of providing a causal explanation of a PDDL plan within the context of the domain and the problem with rigorous mathematical structure. In particular, the structure permits reasoning about compositionality of plans whether it be within the same solution or a transferred solution. In addition, the representation provides a diagrammatic syntax that identifies rectangles as actions and curvy lines, or strings, as literals. The placement of the rectangles and strings relay a temporal progression, read from top to bottom, which allows user to sense the use of literals as the plan progresses. This can lead to insights about action dependencies. Extensions of this work include using other category theory mechanics to describe the maps between plans and human conceptual models to provide a framework to compare solutions across planning domains. 


\bibliography{ref}
\bibliographystyle{named}
\end{document}